# Application of Wrench based Feasibility Analysis to the Online Trajectory Optimization of Legged Robots

Romeo Orsolino[1], Michele Focchi[1], Carlos Mastalli[1,2], Hongkai Dai[3], Darwin G. Caldwell[1] and Claudio Semini[1]

*Abstract*—Motion planning in multi-contact scenarios has recently gathered interest within the legged robotics community, however actuator force/torque limits are rarely considered. We believe that these limits gain paramount importance when the complexity of the terrains to be traversed increases. We build on previous research from the field of robotic grasping to propose two new six-dimensional bounded polytopes named the Actuation Wrench Polytope (AWP) and the Feasible Wrench Polytope (FWP). We define the AWP as the set of all the wrenches that a robot can generate while considering its actuation limits. This considers the admissible contact forces that the robot can generate given its current configuration and actuation capabilities. The Contact Wrench Cone (CWC) instead includes features of the environment such as the contact normal or the friction coefficient. The intersection of the AWP and of the CWC results in a convex polytope, the FWP, which turns out to be more descriptive of the real robot capabilities than existing simplified models, while maintaining the same compact representation. We explain how to efficiently compute the vertex-description of the FWP that is then used to evaluate a *feasibility factor* that we adapted from the field of robotic grasping [1]. This allows us to optimize for robustness to external disturbance wrenches. Based on this, we present an implementation of a motion planner for our quadruped robot HyQ that provides *online* Center of Mass (CoM) trajectories that are guaranteed to be statically stable and actuation-consistent.

## I. INTRODUCTION

LEGGED locomotion in rough terrains requires the careful selection of a contact sequence along with a feasible motion of the CoM. In case of an unexpected event (e.g. changes in the terrain conditions, human operator commands, external force disturbance, inaccuracies in the state estimation and in the terrain mapping, etc.) replanning is an important feature to avoid accumulation of errors. As a consequence, ideal motion planners for complex terrains should be fast but accurate. Approaches that use simplified dynamic models are especially fast but they only capture the main dynamics of the system [2]. On the other hand, other approaches use the whole-body model of the robot and provide particularly accurate joint torques and position trajectories, but are not suitable for online applications in arbitrary terrains. A third option consists in offline learning primitives and behaviors generated with the more accurate whole-body models that can be later quickly realized in real-time [3].

The present paper tackles this issue using simplified dynamic models that still contain sufficient details of the system. The use of the centroidal dynamics [4] coupled with the CWC-based planning represents a step in this direction, allowing to remove the limitation of having coplanar contacts (as for Zero Moment Point (ZMP) based approaches) and thus increasing the complexity of motions that can be generated [5], [6]. This has also led to the formulation of algorithms that can efficiently verify robots stability in multi-contact scenarios [7], [8]. Such approaches however still fail to capture some properties of the robot - such as the actuation limits, the joints kinematic limits and the possible self-collisions. These properties become more and more important with the increasing complexity of the environment and we believe that they should not be neglected in motion planning. To the best of the author's knowledge, while actuation constraints have been considered at the control level [9], [10], this is the first time that a framework for the formulation of actuation consistent online motion planners is provided. As later explained, the strategy consists in devising CoM trajectories that are guaranteed to respect the actuation and friction constraints, without explicitly optimizing neither the joint torques nor the contact forces.

### A. Contribution

In this paper we address the problem of devising actuation-consistent motions for legged robots and, in particular, we propose the four following contributions: a) first, we introduce the concept of Actuation Wrench Polytope (AWP) which complements the CWC, adding the robot-related constraints such as its configuration and actuation capabilities. The consideration of both the environment-related constraints (the CWC) and the robot-related constraints (the AWP) leads to the definition of a second convex polytope that we call Feasible Wrench Polytope (FWP). This can be seen as a development of the Grasp Wrench Space (GWS) proposed for robotic grasping [1]. Disregarding the constraints due to self-collision and kinematic joints limits, the FWP can then be used as a sufficient criterion for legged robots stability; b) second, we exploit recent advancements in computational geometry [11] to compute the vertex-description ($\mathcal{V}$-description) of the FWP, drastically reducing the computation time with respect to the double-description ($\mathcal{D}$-description) based methods; c) third, we adapt the vertex-based *feasibility factor*, as in [1],

Manuscript received December 19, 2017; Revised March, 30, 2018; Accepted April, 22, 2018.
This paper was recommended for publication by Editor Paolo Rocco upon evaluation of the Associate Editor and Reviewers' comments. This work was supported by Istituto Italiano di Tecnologia.
[1]Department of Advanced Robotics, Istituto Italiano di Tecnologia, Genova, Italy. email: {romeo.orsolino, michele.focchi, darwin.caldwell, claudio.semini}@iit.it
[2]CNRS-LAAS, University of Toulouse, Toulouse, France email: carlos.mastalli@laas.fr
[3]Toyota Research Institute (TRI), Los Altos, USA email: hongkai.dai@tri.global
Digital Object Identifier (DOI): see top of this page

to evaluate *online* the feasibility of a motion plan for legged robots. d) Finally, we exploit this feasibility factor for the *online* generation of CoM trajectories that are statically stable and actuation-consistent.

*B. Outline*

The rest of this paper is organized as follows: we first discuss the previous research in the field of wrench based feasibility analysis with a special consideration for the robot stability and actuation-consistency (Section II). We then introduce the computation of the AWP [12] and an efficient strategy to calculate the $\mathcal{V}$-description of the FWP (Section III). Section IV introduces the FWP-based *feasibility metric* and Section V describes how this can be used for *online* motion planning. Section VI presents the simulations and experimental results we obtained by implementing our strategy on the Hydraulically actuated Quadruped (HyQ) robot [13]. Finally, Section VII draws the conclusions with a brief discussion on the results and on future developments.

## II. RELATED WORK

Wrench based feasibility analysis is not a novel idea in robotics. In the field of Cable-Driven Parallel Robots (CDPR) the set of all the configurations that can be realized respecting the maximum tension in the ropes is indicated under the name of Wrench-Feasible Workspace (WFW) [14], [15]. The WFW is used to analyze the robot's capability to carry loads, but it does not consider constraints that might arise from the interaction with the environment, such as unilaterality and friction. The idea of modeling the wrench admissible region is also present in the field of mechanical fixtures and tolerance analysis [16] where reciprocity of twists and screws is exploited to characterize the mobility conditions of any couple of faces in tolerance chains.

In the robotic grasping community it is common to consider sets of wrenches respecting frictional constraints [17]. When considered, actuation limits take the form of an upper bound on the magnitude of the normal contact force. Composing the contribution of each contact friction cone, a GWS can be defined [1], representing a subset of the task wrench space in which the a robust grasp against external disturbances is guaranteed. Such actuation constraints, however, may depend on the joint configuration but they usually disregard the fact that the maximal normal force cannot be coupled with any tangential force component.

In legged locomotion the seminal work of Takao et al. has studied the problem of finding Feasible Solution of Wrench (FSW) in multi-contact configurations [18]. Wrench sets have appeared with the CWC-based margin [5], which is a stability criterion for locomotion that is suitable for *non-coplanar* contacts and finite friction coefficients. Dai et al. in [19], [6] have shown how to exploit a CWC margin to obtain a convex optimization formulation that can plan CoM and joints trajectories of legged robots on complex terrains. On a similar line, Caron et al. [20] have focused on improving the real-time performances of 3D motion planning, either exploiting the double-description of the 6D polyhedra or by considering lower dimensional projections of the CWC defining full-support areas. The latter, coupled with a linear pendulum model, led to the definition of the *pendular support area* [21]. Despite the excellent results shown in this field, the lack of successful experimental implementations on the hardware is mainly due to the fact that often the desired complex movements require the torques to be beyond the limits of the actuators. Indeed, the actuation capabilities become even more critical when the robot interacts with an environment of complex geometry. Therefore, an accurate evaluation of the robot actuation capabilities takes on paramount importance.

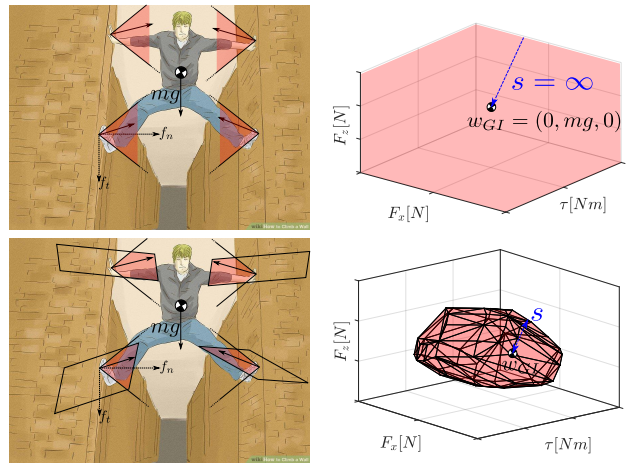

Fig. 1. Planar example: unbounded friction cones (*top-left*) give origin to unbounded feasible wrench sets (*top-right*). Bounded friction polytopes (*bottom-left*), instead, generate bounded feasible wrench sets (*bottom-right*).

## III. WRENCH-BASED ANALYSIS

Considering the actuation limits significantly affects the wrench margin of a legged robot. As an example let us consider a human(oid) trying to climb a vertical chimney shown in Fig. 1. Here, the CWC is obtained through the Minkowski sum of the friction cones, represented by the pink areas (Fig. 1 *top-left*). In the CWC-based approach the margin is quantified as the minimal distance between the gravito-inertial wrench $\mathbf{w}_{GI}$ and the boundary of the CWC. The CWC margin represents the maximum allowed wrench that can be applied (or rejected in case of a disturbance) in order to keep the system stable. The CWC margin $s$ has an infinite value $s = \infty$, i.e. the *force closure* condition is achieved (Fig. 1 *top-right*). This happens because the friction cone representation assumes that a contact force with an infinite normal component can be realized at the contact. This misleading result is the consequence of not taking the actuation limits into account.

On the other hand, imposing the actuation limits can be rephrased as adding a further constraint on the magnitude of the admissible contact forces with a *force polytope* that depends on the actuation capabilities and on the current configuration Fig. 1(*bottom-left*). This limits the set of applicable body wrenches (feasible wrenches) that the human(oid) can apply on its own CoM to keep himself stable. In Fig. 1(*bottom-right*) the bounded volume represents the result of the Minkowski sum of the four *bounded friction polytopes* after considering the torque contribution of each maximal contact force. This convex region is a subset of the CWC and we call it the Feasible Wrench Polytope (FWP). This is computed as the intersection of CWC and AWP, for more details see Section III-B. According to our definition, the margin $s$ is limited to a finite value, as in Fig. 1(*bottom-right*), showing that the human(oid) might fall if his limbs are not strong enough to support his body's weight.

### A. The Actuation Wrench Polytope (AWP)

In this section we illustrate the procedure to compute the AWP, the wrench polytope devoted to taking actuation limits into account. Let us consider the Equation of Motion (EoM) of a floating-base robot with $n_l$ branches (e.g. legs) in contact with the environment, each of them with a number $n_a$ of actuated Degree of Freedoms (DoFs), $n = \sum_{k=1}^{n_l} n_a^k$ being the total number of actuated joints:

$$\mathbf{M}(\mathbf{q})\ddot{\mathbf{q}} + \mathbf{c}(\mathbf{q},\dot{\mathbf{q}}) + \mathbf{g}(\mathbf{q}) = \mathbf{B}\boldsymbol{\tau} + \mathbf{J}_s^T(\mathbf{q})\mathbf{f} \quad (1)$$

where $\mathbf{q} = \begin{bmatrix}\mathbf{q}_b^T & \mathbf{q}_j^T\end{bmatrix}^T \in SE(3) \times \mathbb{R}^n$ represents the pose of the floating-base system, composed of the pose of the base-frame $\mathbf{q}_b \in SE(3)$ and of the generalized coordinates $\mathbf{q}_j \in \mathbb{R}^n$ describing the positions of the $n$ actuated joints. The vector $\dot{\mathbf{q}} = \begin{bmatrix}\mathbf{v}^T & \dot{\mathbf{q}}_j^T\end{bmatrix}^T \in \mathbb{R}^{n+6}$ is the generalized velocity, $\boldsymbol{\tau} \in \mathbb{R}^n$ is the vector of actuated joint torques while $\mathbf{c}(\mathbf{q})$ and $\mathbf{g}(\mathbf{q}) \in \mathbb{R}^{6+n}$ are the centrifugal/Coriolis and gravity terms, respectively. $\mathbf{B} \in \mathbb{R}^{(6+n) \times n}$ is the matrix that selects the actuated joints of the system. $\mathbf{f} \in \mathbb{R}^{3n_l}$ is the vector of contact forces[1] that are mapped into joint torques through the stack of Jacobians $\mathbf{J}_s(\mathbf{q}) \in \mathbb{R}^{3n_l \times (6+n)}$. If we split (1) into its underactuated and actuated parts, we get:

$$\underbrace{\begin{bmatrix}\mathbf{M}_b & \mathbf{M}_{bj} \\ \mathbf{M}_{bj}^T & \mathbf{M}_j\end{bmatrix}}_{\mathbf{M}(\mathbf{q})} \underbrace{\begin{bmatrix}\dot{\mathbf{v}} \\ \ddot{\mathbf{q}}_j\end{bmatrix}}_{\ddot{\mathbf{q}}} + \underbrace{\begin{bmatrix}\mathbf{c}_b \\ \mathbf{c}_j\end{bmatrix}}_{\mathbf{c}(\mathbf{q},\dot{\mathbf{q}})} + \underbrace{\begin{bmatrix}\mathbf{g}_b \\ \mathbf{g}_j\end{bmatrix}}_{\mathbf{g}(\mathbf{q})} = \underbrace{\begin{bmatrix}\mathbf{0}_{6 \times n} \\ \mathbf{I}_{n \times n}\end{bmatrix}}_{\mathbf{B}} \boldsymbol{\tau} + \underbrace{\begin{bmatrix}\mathbf{J}_{sb}^T \\ \mathbf{J}_{sq}^T\end{bmatrix}}_{\mathbf{J}_s(\mathbf{q})^T} \mathbf{f}. \quad (2)$$

By inspecting the actuated part ($n$ bottom equations), resulting from the concatenation of the equations of motions of all the branches, we see that $\mathbf{J}_{sq} \in \mathbb{R}^{3n_l \times n}$ is block diagonal and we can use it to map joint torques into contact forces for each leg *separately*. We will see in Section V-A that this is convenient because it avoids using the coupling term $\mathbf{J}_{sb}$.

For motion planning we are interested in estimating the maximum $\mathbf{f}^{\max} \in \mathbb{R}^{3n_l}$ and minimum $\mathbf{f}^{\min} \in \mathbb{R}^{3n_l}$ contact forces that the end-effectors are able to apply on the environment. This quantity can be estimated by considering the maximal torques achievable by the actuation system $\boldsymbol{\tau}^{\lim} \in \mathbb{R}^n$ in a given configuration $\mathbf{q}_j \in \mathbb{R}^n$ of the actuated joints. As previously anticipated, we can estimate the maximal and minimal admissible contact force $\mathbf{f}_i^{\lim} \in \mathbb{R}^3$ for each end-effector $i$ separately, by considering a subset of the EoM (represented by the $i$ sub-script) describing the dynamics of the actuacted joints of that specific leg:

$$\mathbf{f}_i^{\lim} = \mathbf{J}_i^{T\#}\big(\underbrace{\mathbf{M}_{bi}^T\dot{\mathbf{v}} + \mathbf{M}_i\ddot{\mathbf{q}}_i + \mathbf{c}_i + \mathbf{g}_i}_{\delta} - \boldsymbol{\tau}_i^{\lim}\big) \quad (3)$$

where $(.)^{\#}$ is the Moore-Penrose pseudoinverse, and $\mathbf{J}_i \in \mathbb{R}^{3 \times n_a}$ is the Jacobian matrix for the $i-th$ foot. In our robot, each leg has three DoFs ($n_a = 3$), thus a simple inversion is sufficient since $\mathbf{J}_i$ is square. Note that, for redundant and for underactuated limbs, the use of the pseudo-inverse may lead to solutions that actually violate the bounds given by the torque limits. In such cases, solving an inequality-constrained quadratic program is a viable solution [22].

[1]Note that our robot has nearly point feet, thus we only consider pure forces at the contact point and no contact torque.

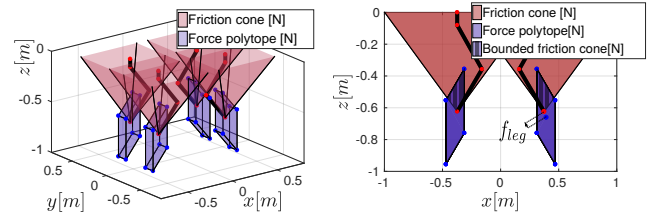

Fig. 2. Representation of the force polytopes (blue) and of the friction cones (pink) of each single leg in 3D (*left*) and projected on the $(F_x, F_z)$ plane (*right*). The offset $f_{leg}$ is due to the bias term $\delta$ in (3).

$\boldsymbol{\tau}_i^{\lim} \in \mathbb{R}^{n_a}$ is a vector that contains the upper and lower bounds of the joint torques of each single leg. Considering all their combinations, these bounds result in $2^{n_a}$ values of $\mathbf{f}_i^{\lim}$ that form the vertices of the *force polytope* $\mathcal{F}_k$. In the case of our quadruped robot HyQ, $\mathcal{F}_k$ is a polytope with 8 vertices and its shape changes nonlinearly with the joint configuration because of the nonlinearities in $\mathbf{J}_i{}^2$. As an example, we compute the force polytope for each leg in a quadruped robot. Fig. 2(*left*) shows the four force polytopes (together with the friction cones) obtained for a typical quadruped robot. Fig. 2(*right*) shows a lateral view that depicts the same force polytopes projected onto the $(F_x, F_z)$ plane.

To compute the AWP, the next step is to add the torque values that are generated in correspondence to the maximum pure contact forces:

$$\mathbf{w}_{i,k} = \begin{bmatrix}\mathbf{f}_{i,k}^{\lim} \\ \mathbf{p}_i \times \mathbf{f}_{i,k}^{\lim}\end{bmatrix} \quad \text{with } k = 1, \ldots, 2^{n_a} \quad (4)$$

where $\mathbf{p}_i \in \mathbb{R}^3$ represents the position of the $i-th$ foot and $\mathbf{w}_{i,k} \in \mathbb{R}^6$ represents the wrench that can be realized at that foot, both quantities are expressed in a fixed frame. Therefore, the set of admissible wrenches that can be applied at the CoM by the $i-th$ foot/end-effector is:

$$\mathcal{W}_i = ConvHull(\mathbf{w}_{i,1}, \ldots, \mathbf{w}_{i,2^{n_a}}) \quad (5)$$

with $i = 1, \ldots, n_l$. We now have $n_l$ wrench polytopes $\mathcal{W}_i$, one for each limb in contact with the environment. Finally, the AWP corresponds to the Minkowski sum of all the $n_l$ wrench polytopes:

$$AWP = \oplus_{i=1}^{n_l} \mathcal{W}_i \quad (6)$$

As defined above, the AWP is a bounded convex polytope in $\mathbb{R}^6$ (Fig. 3 *left*) that contains all the admissible wrenches that can be applied to the robot's CoM that do not violate the actuation limits of the limbs in contact with the environment. Note that the force polytopes $\mathcal{F}_k$ are not zonotopes because they result from the sum of a zonotopic term $-\mathbf{J}_i^{T\#}\boldsymbol{\tau}^{\lim}$ and of a nonzero singleton $\mathbf{J}_i^{T\#}\delta$ as in Eq. 3. This differs from what is generally done for CDPR and for the GWS, where inertial and Coriolis effects are usually ignored and the ropes/fingers are considered light-weighted [23].

[2]The torque limits $\boldsymbol{\tau}_i^{\lim}$ may depend on the joint positions, making the dependency of the polytope from the joints configuration even more complex (e.g. a revolute joint made of a linear actuator with nonlinear lever-arm).

*B. The Feasible Wrench Polytope (FWP)*

Note that the AWP does not include the constraints imposed by the environment, namely, the terrain normal, the friction coefficient and the unilateral contact condition (e.g. the legs can not pull on the ground). However, those constraints can be accounted by the CWC [5] (Fig. 3 *center*):

$$CWC = ConvexCone(\hat{\mathbf{e}}_i^k) \quad k = 1, \ldots n_e \quad (7)$$

and $i = 1, \ldots n_l$. Here $n_e$ is the number of edges of the linearized friction cone and:

$$\hat{\mathbf{e}}_i^k = \begin{bmatrix} \mathbf{e}_i^k \\ \mathbf{p}_i \times \mathbf{e}_i^k \end{bmatrix} \in \mathbb{R}^6, \quad \text{with} \quad k = 1, \ldots, n_e \quad (8)$$

where $\mathbf{e}_i^k \in \mathbb{R}^3$ is the $k-th$ edge of the contact point $i$. We subsequently perform the intersection of the CWC with the AWP obtaining a convex polytope that we define as Feasible Wrench Polytope (FWP) (Fig. 3 *right*):

$$FWP = CWC \cap AWP. \quad (9)$$

However, performing the *intersection* of polytopes in 6D is an expensive operation that requires the $\mathcal{D}$-description [11] of both operands. We propose a more efficient approach for the computation of the FWP that: 1) first computes the intersection between the friction cones $\mathcal{C}_i$ and the force polytopes $\mathcal{F}_i$ obtaining, for each $i-th$ contact, a 3D bounded friction cone $\mathcal{B}_i$ (Fig. 2 *right*) with $\mu$ vertices $\mathbf{b}_i^k \in \mathbb{R}^3$:

$$\mathcal{B}_i = ConvHull(\mathbf{b}_i^k), \quad \text{with} \quad k = 1, \ldots, \mu \quad (10)$$

2) then composes the wrench by adding the torque, as in (4):

$$\hat{\mathbf{b}}_i^k = \begin{bmatrix} \mathbf{b}_i^k \\ \mathbf{p}_i \times \mathbf{b}_i^k \end{bmatrix} \in \mathbb{R}^6 \quad \text{with} \quad k = 1, \ldots, \mu \quad (11)$$

obtaining in this way the intermediate sets $\mathcal{G}_i \in \mathbb{R}^6$:

$$\mathcal{G}_i = ConvHull(\hat{\mathbf{b}}_i^k), \quad \text{with} \quad k = 1, \ldots, \mu \quad (12)$$

3) finally, the FWP is computed through the Minkowski sum of the $\mathcal{G}_i$ of all the $n_l$ contacts:

$$FWP = \oplus_{i=1}^{n_l} \mathcal{G}_i \quad (13)$$

The advantage of this proposed method is that the intersection is performed in 3D rather than in 6D, which is computationally faster. This is advantageous also for the final step in (13) because it avoids computing vertices that will be removed later (e.g. all the vertices from the AWP with negative contact forces are removed by intersecting with the CWC). Additionally, the Minkowski sum can be efficiently obtained using the $\mathcal{V}$-description only as in [11].

*C. Polytope representation for a Planar model*

To achieve a better understanding of the nature of these polytopes, let us consider the simplified case of a planar dynamic model, as in Fig. 2 (*right*), where each point of the space is represented through the $(x, z)$ coordinates. In this case the wrench space has three coordinates $(F_x, F_z, \tau_y)$ and can be represented in 3D. Fig. 3 depicts the AWP (*left*), the CWC (*center*) and the FWP (*right*) for this simplified model.

## IV. FWP-BASED FEASIBILITY METRIC

The CWC margin has been proven to be a universal criterion for dynamic legged stability [5]. However, the CWC still lacks knowledge of robot's feasibility constraints such as self-collisions, actuation and kinematic joint limits. These constraints become even more important when the roughness of the terrain increases. Therefore, in order to plan complex motions in unstructured environments we will introduce a more restrictive metric, that we generically call the *feasibility metric*. This metric incorporates all the properties of the CWC criterion, and additionally the robot's actuation limits.

In order to obtain the *feasibility metric* we first need to compute the robot's gravito-inertial wrench $\mathbf{w}_{GI} \in \mathbb{R}^6$ in the specific robot state that we want to evaluate:

$$\mathbf{w}_{GI} = \dot{\mathbf{h}} - \mathbf{w}_G \quad (14)$$

with:

$$\dot{\mathbf{h}} = \begin{bmatrix} m\ddot{\mathbf{c}} \\ \dot{\mathbf{k}}_\mathcal{W} \end{bmatrix}, \quad \mathbf{w}_G = \begin{bmatrix} m\mathbf{g} \\ \mathbf{c} \times m\mathbf{g} \end{bmatrix} \quad (15)$$

where $\mathbf{k}_\mathcal{W} \in \mathbb{R}^3$ is the robot's angular momentum and $\mathbf{c}$ is its CoM position (both expressed in the fixed coordinate frame $\mathcal{W}$). The criterion of feasibility can then be defined as:

$$\mathbf{w}_{GI} \in FWP. \quad (16)$$

The definition of feasibility metric depends on the type of the representation chosen for the FWP, i.e. half-plane description ($\mathcal{H}$-description) or a vertex description ($\mathcal{V}$-description).

*A. Half-plane description*

In the $\mathcal{H}$-representation, the FWP set can be written in terms of half-spaces as:

$$FWP = \{\mathbf{w} \in \mathbb{R}^6 | \hat{\mathbf{a}}_j^T \mathbf{w} \leq \mathbf{0}, j = 1, \ldots n_h\} \quad (17)$$

where $n_h$ is the number of half-spaces of the FWP and $\hat{\mathbf{a}}_j \in \mathbb{R}^6$ is the normal vector to the $j-th$ facet. The feasibility criterion expressed in (16) can thus be written as:

$$\mathbf{H}^T \mathbf{w}_{GI} \leq \mathbf{0} \quad (18)$$

where $\mathbf{H} \in \mathbb{R}^{6 \times n_h}$ is the matrix whose columns are the normals to all the half-spaces of the FWP and $\leq$ is a component-wise operator. The columns of $\mathbf{H}$ can be divided into two blocks $\mathbf{H}_c$ and $\mathbf{H}_a$ in order to differentiate the CWC half-spaces from the AWP half-spaces, respectively: $\mathbf{H} = [\mathbf{H}_c | \mathbf{H}_a]$. If $\mathbf{H}_c^T \mathbf{w}_{GI} > \mathbf{0}$ but $\mathbf{H}_a^T \mathbf{w}_{GI} \leq \mathbf{0}$ then the robot's state is consistent with its actuation capabilities but its contact condition is unstable (e.g. friction limits are violated). Viceversa, if $\mathbf{H}_c^T \mathbf{w}_{GI} \leq \mathbf{0}$ but $\mathbf{H}_a^T \mathbf{w}_{GI} > \mathbf{0}$ then the system has stable contacts but it does not respect the actuation limits. In the latter case, the legged system might still not fall but it will not be able to realize the desired task.

If the $\mathcal{H}$-description of the FWP is given, we can provide a definition of robustness that extends the properties of the CWC margin. In the same line with [6] the feasibility metric can be defined as the margin $m$, i.e. the distance, of the point $w_{GI}$ from the boundaries of the FWP. This corresponds to finding the biggest disturbance wrench $\mathbf{w}_d \in \mathbb{R}^6$ that the system can reject. This is equivalent to computing the largest residual radius $m$ such that the $m$-ball $\mathcal{B}_m$ (centered in $\mathbf{w}_{GI}$) still lies within the FWP:

$$\mathcal{B}_m \in FWP \quad (19)$$

where $\mathcal{B}_m$ is defined as:

$$\mathcal{B}_m = \{\mathbf{w}_{GI} + \mathbf{T}(\mathbf{p}_{w_d})\mathbf{w}_d \mid \mathbf{w}_d^T \mathbf{Q} \mathbf{w}_d \leq m\} \quad (20)$$

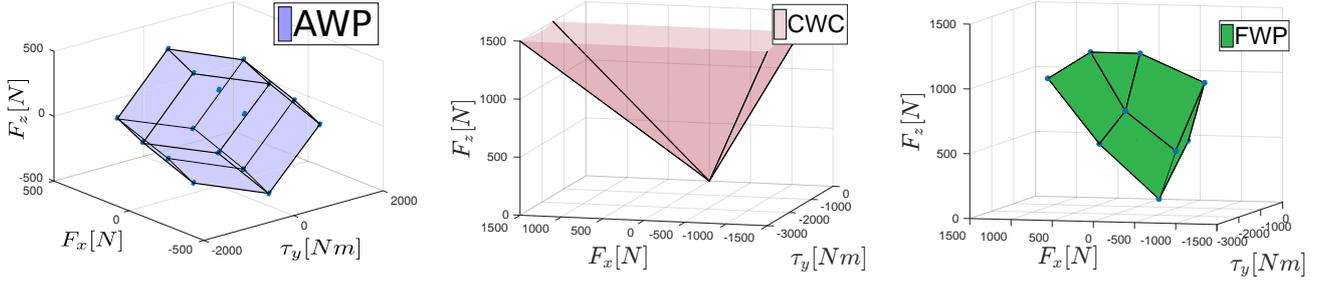

Fig. 3. The Actuation Wrench Polytope (AWP) (*left*), the Contact Wrench Cone (CWC) (*center*) and the Feasible Wrench Polytope (FWP) (*right*). These drawings refer to a planar dynamic model where the only non zero wrench components are $(F_x, F_z, \tau_y)$ and that can therefore be represented in 3D.

$\mathbf{p}_{w_d}$ is the disturbance application point and $\mathbf{T}(\mathbf{p}_{w_d})$ is the adjoint spatial transform that expresses it in the frame $\mathcal{W}$ [6]. $\mathbf{Q}$ is a positive definite matrix that is used to make the units of the wrench homogeneous.

### B. Vertex description

If only a $\mathcal{V}$-description of the FWP is available, the distance between $w_{GI}$ and the faces of the FWP cannot be computed anymore and a different definition of feasibility metrics is needed. We decide to employ in this case a feasibility scalar factor $s \in (-\infty, 1]$ adapted by the scaling factor defined in [1] used to measure a robotic grasp quality.
Let us consider a matrix $\mathbf{V} \in \mathbb{R}^{6 \times n_v}$ whose columns are the vertices $\mathbf{v}_i$ of the FWP, and a vector $\boldsymbol{\lambda} \in \mathbb{R}^{n_v}_+$ of non-negative weights where $n_v$ is the number of vertices of the FWP. Every point inside the FWP can be described with a combination of weights $\lambda_i$ such that $\sum_{i=0}^{n_v} \lambda_i = 1$. We therefore define the robot to be in a feasible state if, for the corresponding wrench $\mathbf{w}_{GI}$, there exists a $\boldsymbol{\lambda}$ such that $\mathbf{V}\boldsymbol{\lambda} = \mathbf{w}_{GI}$, with $\|\boldsymbol{\lambda}\|_1 = 1$ and $\lambda_i \geq 0 \quad i = 1, \ldots, n_v$.
A preliminary step for the computation of the feasibility factor consists in subtracting the centroid $\mathbf{v}_c$ from all the FWP vertices $\mathbf{v}_i$, obtaining new translated vertices $\hat{\mathbf{v}}_i$ ($\hat{\mathbf{v}}_i = \mathbf{v}_i - \mathbf{v}_c$). This has the effect of shifting the origin of the wrench space in the centroid, that, in a $\mathcal{V}$-representation, is a good approximation of the most "robust" point (e.g. the Chebishev centre). We then define a new *shrunk* polytope $\mathcal{P}^s$ centered in the origin (which is now also the centroid of the FWP). $\mathcal{P}^s$ can be expressed in terms of its own vertices $\hat{\mathbf{v}}_i^s$ and of a set of multipliers $\boldsymbol{\lambda}_i^s$:

$$\mathcal{P}^s = \left\{\mathbf{w} \in \mathbb{R}^6 | \mathbf{w} = \sum_{i=1}^{n_v} \lambda_i^s \hat{\mathbf{v}}_i^s, \lambda_i^s \geq 0, \|\boldsymbol{\lambda}^s\|_1 = 1\right\} \quad (21)$$

For a better understanding Fig. 4 (*left*) illustrates the idea of the shrunk polytope for a 2D representation.
The FWP's vertices $\hat{\mathbf{v}}_i$ are linked to the vertices of $\mathcal{P}^s$ through the feasibility factor $s$:

$$\hat{\mathbf{v}}_i^s = \hat{\mathbf{v}}_i(1-s), \quad -\infty < s \leq 1 \quad (22)$$

If, for instance, $s = 1$ then $\mathcal{P}^s$ shrinks into the origin. If we impose $\lambda_i = (1-s)\lambda_i^s$, then we can write the shrunk polytope $\mathcal{P}^s$ in terms of the vertices $\hat{\mathbf{v}}_i$ of the FWP, i.e.:

$$\mathcal{P}^s = \left\{\mathbf{w} \in \mathbb{R}^6 | \mathbf{w} = \sum_{i=1}^{n_v} \lambda_i \hat{\mathbf{v}}_i, \lambda_i \geq 0, \|\boldsymbol{\lambda}\|_1 = 1-s\right\} \quad (23)$$

We can see the feasibility factor as the scalar $s$ that corresponds to the smallest shrunk polytope containing the point $\mathbf{w}_{GI}$. The problem of finding $s$ can be formulated as a Linear Program (LP) that can be carried out by any general-purpose solver:

$$\begin{aligned}\max_{\boldsymbol{\lambda}, s} \quad & s \\ \text{s.t.} \quad & \mathbf{V}\boldsymbol{\lambda} = \mathbf{w}_{GI} \\ & \|\boldsymbol{\lambda}\|_1 = 1-s \quad s \in (-\infty, 1] \\ & \lambda_i \geq 0 \quad i = 1, \ldots, n_v\end{aligned} \quad (24)$$

Note that the larger is $s$, the more robust is the system against disturbances. A negative $s$ means that the point is out of the polytope and the wrench is unfeasible. When $s$ becomes zero, it means the point is on the polytope boundary and that either the friction or actuation limits are violated. Table I shows the computation time of a Intel(R) Core(TM) i5-4440 CPU @ 3.10GHz with 4 cores for three and four contacts scenarios. The feasibility factor $s$, unlike the margin $m$, is not sensitive to the fact of having different units in the wrench space since it does not encode the concept of distance. On the other hand, for the very same reason, the factor $s$ of a given joint configuration cannot be compared to another configuration, because of the different scaling of the two polytopes.

## V. ONLINE TRAJECTORY OPTIMIZATION (TO)

The FWP factor can be used to devise a motion planner that provides robust CoM trajectories. Hereafter, we present a brief description of how we used the proposed criterion to plan

TABLE I
COMPUTATION TIME OF THE FEASIBILITY FACTOR $s$

|  | 3 non-coplanar contacts | 4 non-coplanar contacts |
|---|---|---|
| FWP vertices | 436 | 1118 |
| variables | 437 | 1119 |
| constraints | 7 | 7 |
| LP time [ms] | 90 | 350 |

online motions for our quadruped robot, that do not violate the actuation limits.

We further extended the capabilities of the locomotion framework [24], [25] by replacing the original (*heuristic*) planner with a trajectory optimizer that exploits the proposed feasibility criterion. This framework realizes a statically stable *crawling* gait where a *base motion* phase and a *swing motion* phase are alternated (therefore the base of the robot does not move when a leg is in swing).

To be able to compare with the heuristic planner, we opted for a decoupled planning approach where the footholds and the CoM trajectory of the robot are determined sequentially.

Our online TO computes during every swing phase the CoM trajectory to be realized in the next base motion phase using a *one-step horizon*. The decision variables of the optimization problem are the $X$, $Y$ components of the CoM positions, the velocities and the duration of the *base motion* phase $\Delta t_{bm}$: $\mathbf{\Gamma} = \{\mathbf{c}_x[k], \mathbf{c}_y[k], \dot{\mathbf{c}}_x[k], \dot{\mathbf{c}}_y[k], \Delta t_{bm}\}$ with $k = 1, \cdots, N$. The trajectory is discretized in $N$ equally spaced knots (at time intervals $h = \Delta t_{bm}/N$). Note, that we here do not optimize the angular dynamics nor the coordinate $z$, parallel to gravity. This is because, for quasi static motions, the predominant acceleration term acting on the system is gravity itself, and therefore its influence on the stability or on the joint torques is limited compared to the role of the $X$ and $Y$ components. We aim to maximize the FWP factor $s$, as in Eq. 24, while we enforce back-ward Euler integration constraints along the trajectory and zero velocity at the trajectory extremes:

$$\min_{\mathbf{\Gamma}} \sum_{k=1}^{N} \mathcal{L}(\mathbf{c}[k], \dot{\mathbf{h}}[k], \mathbf{V})$$
$$\text{s.t.} \quad \dot{\mathbf{c}}_x[k+1] = (\mathbf{c}_x[k+1] - \mathbf{c}_x[k])/h \quad (25)$$
$$\dot{\mathbf{c}}_y[k+1] = (\mathbf{c}_y[k+1] - \mathbf{c}_y[k])/h$$
$$\dot{\mathbf{c}}_x[0] = \dot{\mathbf{c}}_y[0] = \dot{\mathbf{c}}_x[N] = \dot{\mathbf{c}}_y[N] = \mathbf{0}$$

As a first step, we evaluate the FWP polytope considering the robot contact configuration. By assuming a quasi-static condition ($\ddot{\mathbf{q}} = \dot{\mathbf{q}} = \mathbf{0}$), Eq. 3 reduces to:

$$\mathbf{f}_i^{\lim} = \mathbf{J}_i(\mathbf{q}_{0_i})^{-T}\Big(\mathbf{g}(\mathbf{q}_{0_i}) - \mathbf{B}\boldsymbol{\tau}^{\lim}(\mathbf{q}_{0_i})\Big) \quad (26)$$

Exploiting the approximation later explained in Section V-A we compute the FWP just once at the beginning of the optimization. Then, to compute the running cost $\sum_{k=1}^{N} \mathcal{L}$, for each optimization loop, we evaluate the CoM acceleration along the trajectory ($h \cdot \ddot{\mathbf{c}}_{x,y}[k+1] = \dot{\mathbf{c}}_{x,y}[k+1] - \dot{\mathbf{c}}_{x,y}[k], \forall k$) and evaluate the gravito-inertial wrench at each knot through (14). In order to exploit the $\mathcal{V}$-description, for each node, we should add the $\boldsymbol{\lambda}$ vector as decision variable and the constraints in (24). However, the amount of decision variables would significantly increase due to the high number of vertices in the polytope (i.e. $\boldsymbol{\lambda}$ may have hundreds of elements for each optimization knot) leading to computation times that do not meet the requirements for online planning. Thus we tackled this problem by computing the set of $\boldsymbol{\lambda}$ through a simple Moore-Penrose pseudo-inversion: $\boldsymbol{\lambda}[k] = \mathbf{V}^{\#}\mathbf{w}_{GI}[k]$. In this way the decision variables will be only the states $\mathbf{\Gamma}$ of the system and the number of vertices will influence the size of the TO problem only marginally (see Table III). We noticed that adding a *bias* term in the nullspace of $\mathbf{V}$ to "drive" the solution $\boldsymbol{\lambda}[k]$ toward $\boldsymbol{\lambda}_0 \in \mathbb{R}^{n_v}$ was giving satisfactory results:

$$\boldsymbol{\lambda}[k] = \mathbf{V}^{\#}\mathbf{w}_{GI}[k] + \mathbf{N}_V \boldsymbol{\lambda}_0, \quad \boldsymbol{\lambda} \in \mathbb{R}^{n_v} \quad (27)$$

where $\mathbf{N}_V \in \mathbb{R}^{n_v \times n_v}$ is the null-space projector associated to $\mathbf{V}$. Indeed, if we set $\boldsymbol{\lambda}_0 = [1/n_v, \cdots, 1/n_v]$ as the geometric center of the FWP, the constraints $\|\boldsymbol{\lambda}_0\|_1 = 1, \lambda_{0i} > 0$ are satisfied by construction. Thanks to the one-to-one correspondence between the gravito-inertial wrench and the weights $\boldsymbol{\lambda}$, penalizing the deviation of $\boldsymbol{\lambda} \in \mathbb{R}^{n_v}$ from $\boldsymbol{\lambda}_0$ is equivalent to maximizing the feasibility factor. Therefore, we formulate the running cost computation as:

$$\mathcal{L}\Big[(\mathbf{c}[k], \dot{\mathbf{h}}[k], \mathbf{V})\Big] = \|\boldsymbol{\lambda}[k] - \boldsymbol{\lambda}_0\|_2 \quad (28)$$

TABLE II
FWP'S $\mathcal{V}$- AND $\mathcal{H}$-DESCRIPTION COMPUTATION TIME WITH POLITOPIX [26].

|  | 2 contact points | 3 contact points | 4 contact points |
|---|---|---|---|
| $\mathcal{V}$-description | 0.03s | 0.15s | 0.49s |
| $\mathcal{H}$-description | 0.04s | 1.03s | 30.21s |

TABLE III
TO OF THE VARIABLES $\mathbf{\Gamma}$ USING THE FWP $\mathcal{V}$-DESCRIPTION

|  | 3 non-coplanar contacts | 4 non-coplanar contacts |
|---|---|---|
| timesteps | 10 | 10 |
| FWP vertices | 436 | 1118 |
| variables $\mathbf{\Gamma}$ | 41 | 41 |
| constraints | 24 | 24 |
| time [ms] | 75 | 85 |

### A. Computational issues and approximations

Despite the remarkable computational speed-up obtained by the use of the $\mathcal{V}$-description (see Table II), the evaluation of the FWP still represents the most time-consuming ($a$) of this pipeline (about $150ms$ needed for a triple-stance configuration). The subsequent step ($b$), i.e. the solution of the TO problem for a given FWP, requires instead about $75 - 85ms$ for 10 nodes trajectory. In theory we should recompute step $a$ and step $b$ iteratively, concurrently optimizing the vertices of the FWP and the trajectory inside it. We decided instead to compute the FWP only once per step, assuming that its vertices do not change during the execution of each phase, as explained in the following paragraph.

We analyze the influence on the estimated maximum and minimum contact force when the Jacobian matrix is approximated to be constant along a *body motion* of the HyQ robot. In Fig. 5 we can see the contact force boundaries ($Z$ component only) when a foot spans its workspace (the foot covers all the $X$ and $Y$ positions on a plane located at $Z = -0.6m$). The green surfaces represent the real boundaries of the vertical contact force considering the correct leg Jacobian and correct piston lever-arm for each considered position. The red surfaces show instead the same force boundaries when the correct piston lever-arm and a constant Jacobian is evaluated. We can see that in a neighborhood of the default foot configuration ($[0.3, 0.2, -0.6]m$ with respect to the base frame of the robot) the approximation is accurate and becomes rough in proximity of the workspace boundaries. We chose to use a constant Jacobian matrix corresponding to a joint configuration $\mathbf{q}_0$ of the trajectory coming from the heuristic plan. In this way the Jacobian remains constant and we remove the AWP's dependency from the joints position.

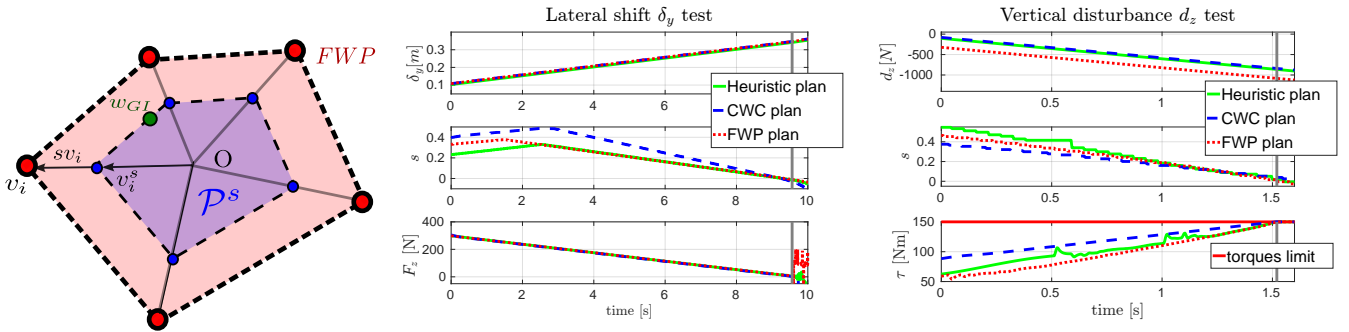

Fig. 4. A 2D pictorial representation of the shrunk polytope $\mathcal{P}^s$ (*left*): we see that the factor $s$ can be seen as the shrinkage rate of the reduced polytope with respect to the FWP; simulation data of the horizontal displacement test (*center*) and vertical disturbance test (*right*).

As a further simplification we assume a quasi-static motion as in Eq. 26. These assumptions allow us to compute the FWP only once at each stance change. Note that all the wrenches are expressed in the fixed frame.

## VI. EXPERIMENTAL RESULTS

In this section we present simulation results and real experiments with the HyQ robot. The first two simulations validate the feasibility factor formulation based on the $\mathcal{V}$-description of the FWP. After that we highlight the differences between our proposed feasibility metric and the state-of-the-art stability metric. Finally we present a few examples of the behaviors we can obtain with the TO presented in Section V.

### A. FWP-based feasibility factor validation

In a first test, we consider three different motion planners: the heuristic planner, a CWC planner (that incorporates only frictional constraints) and our FWP planner. We have the robot crawling where the CoM trajectory is planned by the three different methods and, in all cases, we stop the robot during a triple-stance phase (being more critical for robustness than four-stance phases). We then make the robot displace laterally with an increasing offset $\delta_y = \epsilon 0.5\ m$ ($\epsilon \in \mathbb{R}$ increases linearly from 0 to 1). The objective is to obtain a gradual unloading of the lateral legs and therefore violate the unilaterality constraints. Figure 4 (*center*) shows the evolution of the displacement $\delta_y$ and of the normal component of the contact force $F_z$ at the left-front ($LF$) leg. As expected the plot shows that for all the cases $s$ drops to zero when the leg $LF$ becomes unloaded ($F_z = 0$).

In the second test the robot is again stopped during a walk in a three-legs stance configuration. This time a vertical disturbance force was applied at the origin of the base link (i.e. the geometric center of the torso). The force is *vertical* and pointing downwards with increasing magnitude $d_z = -\epsilon 1000\ N$ where $\epsilon \in \mathbb{R}$ is linearly increasing from 0 to 1. The joint torques will increase because of the action of this force, eventually making one (or more) of them hit the limits. Since the test is performed in a static configuration and the disturbance force is always vertical, the Center of Pressure (CoP) of the system will not change, being the robot always statically stable. Fig. 4 (*right*) shows a plot of the magnitude of the vertical pushing force $d_z$ together with the knee joint torque of the $LF$ leg and the feasibility factors $s$ in the three cases. We can see that, in the case of the static configuration found with the FWP planner, the torque limit is reached for a higher amplitude of the disturbing force (about $-1100N$ compared to $-900N$), showing that this is more robust against external disturbance forces than the configurations selected by the heuristic and by the CWC planners. We can see that in all the cases the feasibility factor $s$ goes to zero when a torque limit is violated.

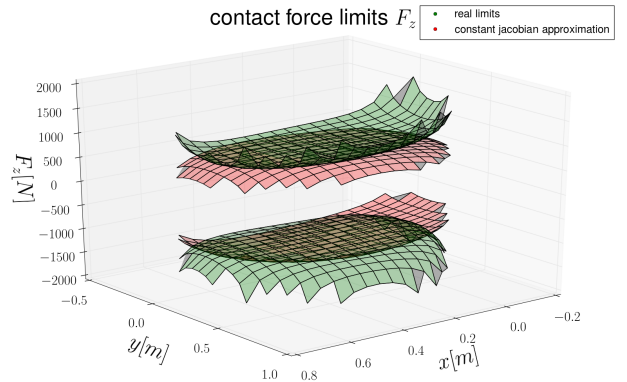

Fig. 5. Contact force limits ($Z$ component) on the left-front (LF) leg of HyQ as a function of the foot position computed with real torque limits (*green*) and with jacobian approximation (*red*).

### B. CWC-margin vs. FWP-factor Comparison

The last test highlights the main differences between the feasibility metrics $s$ and the traditional stability measures. As state-of-the-art stability metric we consider the CWC-margin, which is obtained by applying Eq. 24 on the $\mathcal{V}$-description of the CWC, rather than the FWP as explained in Section IV-B. Fig. 6 (*above*) shows the results when a crawl gait is evaluated using this method. The red line represents the value of the CWC-margin during the triple-stance phase of a crawling gait. The dashed blue line represents the same walk, evaluated again with the CWC-margin, in the case that an external load of $20kg$ is applied on the CoM of the robot during its walk. The factor referring to the four-legs stance is not directly comparable to the triple-stance phase because the vertices of the FWP have a different scaling. For this reason we only show the values referring to the triple stance. We can see that the same two trials, with and without external load, provide a completely different result if evaluated with the FWP-factor (Fig. 6 *bottom*): the blue-dashed line shows that the feasibility is lower in the case with external load.

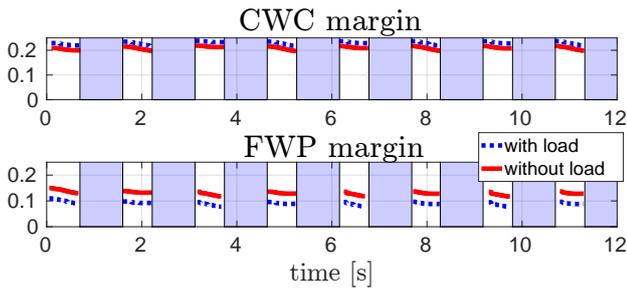

Fig. 6. Evaluation of the same heuristic crawl gait (with and without $20kg$ load) with the CWC margin (*top*) and with the FWP factor (*bottom*).

This implies that, when the load increases, even if the stability might improve, the risk of hitting the torque limits is higher.

*C. Crawling Simulations*

As shown in the accompanying video[3], we report a few simulation and hardware experiments of HyQ performing a crawling gait. At first we see that the heuristic crawl easily hits the torque limits while crawling on a flat ground while carrying an external load of $20kg$ (about 25% of the robot total weight) placed on its CoM. We can then see that the FWP planner, as explained in Section V, finds a new duration $\Delta t_{bm}$ of the *base motion* phase and a new CoM trajectory that avoids hitting the torque limits at all times, while maintaining the desired linear speed.

Final simulations show the capability of the planner to optimize feasible trajectories when the robot has a *hindered* joint (i.e. when a specific joint can only realize a significantly smaller torque than the other joints), or when we limit the normal force that a specific leg can realize on the ground. The video also shows hardware experiments of HyQ crawling on a rough terrain without hitting the torque limits.

## VII. CONCLUSION

The complexity of a motion increases with the complexity of the terrain to be traversed. Moreover, there is a need for online motion replanning to avoid error accumulation. For these reasons, in this paper, we presented the concepts of AWP and FWP and a method to efficiently compute their $\mathcal{V}$-description. We then adapted a *feasibility factor*, originally proposed for grasping [1], to the $\mathcal{V}$-description of the FWP in order to study the stability and the actuation-consistency of a given motion plan. Finally we showed how this factor can be used in a CoM trajectory optimization not only for feasibility evaluation but also for motion planning.

Thanks to the efficiency of the vertex-based approach we are able to perform *online* TO where the robot plans during each *swing* the trajectory for the next *base motion* phase. Our approach does not take any assumption on the environment and it is therefore suitable for complex terrain scenarios.

Future works will concentrate on the removal of the approximations mentioned in Section V-A and on the integration of planned and reactive locomotion behaviors [27].

## ACKNOWLEDGMENT

The authors are grateful to Vincent Delos and Prof. Denis Teissandier of the I2M lab., University of Bordeaux, for the fruitful discussions about computational geometry.

[3] https://youtu.be/vUx5b5kfRfE